\documentclass[conference]{IEEEtran}
\IEEEoverridecommandlockouts
\usepackage{cite}
\usepackage{amsmath,amssymb,amsfonts}
\usepackage{algorithm,algorithmic}
\usepackage{graphicx}
\usepackage{textcomp}
\usepackage{xcolor}
\usepackage{multirow}
\usepackage{booktabs}

\def\BibTeX{{\rm B\kern-.05em{\sc i\kern-.025em b}\kern-.08em
    T\kern-.1667em\lower.7ex\hbox{E}\kern-.125emX}}
\begin{document}

\title{Proto-OOD: Enhancing OOD Object Detection with Prototype Feature Similarity\\

}
% \author{\IEEEauthorblockN{Anonymous Authors}}
\author{\IEEEauthorblockN{Junkun Chen}
\IEEEauthorblockA{\textit{Institute of Computing Technology} \\
\textit{Chinese Academy of Sciences}\\
chenjunkun22s@ict.ac.cn}
\and
\IEEEauthorblockN{Jilin Mei}
\IEEEauthorblockA{\textit{Institute of Computing Technology} \\
\textit{Chinese Academy of Sciences}\\
meijilin@ict.ac.cn}
\\
\and
\IEEEauthorblockN{Liang Chen}
\IEEEauthorblockA{\textit{Institute of Computing Technology} \\
\textit{Chinese Academy of Sciences}\\
chenliang@ict.ac.cn}
\and
\IEEEauthorblockN{Fangzhou Zhao}
\IEEEauthorblockA{\textit{Institute of Computing Technology} \\
\textit{Chinese Academy of Sciences}\\
zhaofangzhou@ict.ac.cn}
\and
\IEEEauthorblockN{Yan Xing}
\IEEEauthorblockA{\textit{Beijing Institute of Control Engineering} \\
xingyan@bice.org.cn}
\and
\IEEEauthorblockN{Yu Hu}
\IEEEauthorblockA{\textit{Institute of Computing Technology} \\
\textit{Chinese Academy of Sciences}\\
huyu@ict.ac.cn}
}

\maketitle

\begin{abstract}
Neural networks that are trained on limited category samples often mispredict out-of-distribution (OOD) objects.
We observe that features of the same category are more tightly clustered in feature space, while those of different categories are more dispersed. Based on this, we propose using prototype similarity for OOD detection. Drawing on widely used prototype features in few-shot learning, we introduce a novel OOD detection network structure (Proto-OOD). Proto-OOD enhances the representativeness of category prototypes using contrastive loss and detects OOD data by evaluating the similarity between input features and category prototypes. During training, Proto-OOD generates OOD samples for training the similarity module with a negative embedding generator. When Pascal VOC are used as the in-distribution dataset and MS-COCO as the OOD dataset, Proto-OOD significantly reduces the FPR (false positive rate). Moreover, considering the limitations of existing evaluation metrics, we propose a more reasonable evaluation protocol.
The code will be released.
\end{abstract}


\section{INTRODUCTION}

AI systems face a significant gap between training and real-world environments. Models trained on limited categories(In-Distribution, ID) encounter unkown category data(Out-Of-Distribution, OOD) in real-world deployment.
Unreliable predictions on OOD data can cause severe accidents, like misidentifying impassable areas in autonomous driving.

Recently, OOD object detection draws research interest\cite{Wilson_2023_ICCV,du2022siren,du2022vos}. The usual methods train a classifier for OOD detection. SAFE\cite{Wilson_2023_ICCV} expands training data with adversarial samples as OOD data, but the similar distribution of their features limits its adaptability to diverse test scenarios. VOS\cite{du2022vos} uses the feature distance to sample outliers, but assessing the optimal distance is a challenge, making it difficult to accurately sample OOD features and train an effective OOD classifier. We believe that the key to solving these issues is extracting highly discriminative features for each category.

The prototypes show strong feature representation capabilities in few-shot learning\cite{snell2017prototypical,DBLP:journals/corr/abs-1911-10714}, inspiring this paper to incorporate prototypes to detect OOD objects. 
As shown in Fig. 1, we observe that features of the same category are more tightly clustered in the feature space, while different categories are more dispersed. 
By collecting features to form prototypes and comparing the similarity between input features and prototypes, we can determine whether the data are OOD objects. High similarity indicates in-distribution data, while low similarity suggests OOD data. Inspired by the relation network\cite{sung2018learning}, we use MLP to estimate the similarity between the input features and the prototypes.
\begin{figure}[htp]
    \includegraphics[scale=0.5]{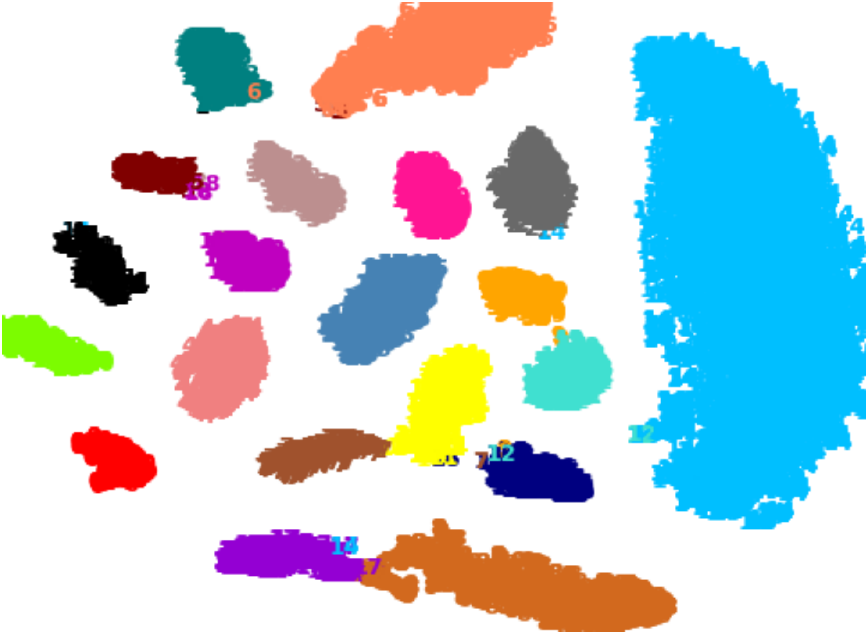}
    \centering
    \caption{The feature visualization of Pascal VOC testing set. We employ t-SNE for dimensionality reduction of the feature extractor's outputs, and each color represents a category feature.}
\end{figure}

This paper introduces Proto-OOD, a novel framework that leverages prototypes to detect OOD objects.
Proto-OOD collects features as prototypes and trains a similarity module to predict the similarity between input features and prototypes. During the training stage, Proto-OOD updates prototypes with a weighting  factor and adds a contrastive loss to make the prototypes more representative. Proto-OOD uses a negative embedding generator to sample OOD features and trains the similarity module with these samples. In the testing stage, Proto-OOD achieves excellent OOD detection performance on the MS-COCO dataset.

Furthermore, our experiments expose the shortcomings of previous evaluation protocols. Specifically, past methods\cite{Wilson_2023_ICCV,du2022siren,du2022vos} don't filter out inaccurate predictions when calculating metrics on ID datasets, leading to inaccurate results. We improve the evaluation protocol and retest the performance of previous methods\cite{Wilson_2023_ICCV,du2022siren,du2022vos} using the updated protocol.

The contributions are summarized as follows:

\textbf{1.} Proto-OOD introduces prototype features into OOD object detection and achieves promising experimental results.

\textbf{2.} Proto-OOD introduces a similarity module and a negative embedding generator. The similarity module calculates the similarity between input features and prototypes for OOD detection. The negative embedding generator produces OOD features for training similarity module, reducing the need for additional OOD samples.

\textbf{3.} Proto-OOD surpasses current methods on MS-COCO\cite{DBLP:journals/corr/LinMBHPRDZ14} and shows significant performance on OpenImages\cite{OpenImages}. Furthermore, we improve the rationality of evaluation protocol by filtering out false positives.
\section{Related Work}
\subsection{OOD Detection for Image Classification}

Researchers propose numerous methods for OOD detection in image classification tasks\cite{abdelzad2019detecting,huang2020feature,zaeemzadeh2021out,wilson2023hyperdimensional,van2020uncertainty,sun2022out,tao2023nonparametric}.
Vahdat et al.\cite{abdelzad2019detecting} suggest that certain optimal network layers in feature space can effectively differentiate ID data from OOD data based on their output features.
Huang et al.\cite{huang2020feature} suggest that OOD data features are clustered in feature space and distinct from ID feature distributions. 
Smith et al.\cite{van2020uncertainty} train a RBF network to identify OOD data by evaluating the uncertainty between input features and centroids. 
In addition, some methods\cite{mahmood2020multiscale,tack2020csi} introduce noise into features to generate OOD samples to train networks.
Mahalanobis distance\cite{lee2018simple,ren2021simple} is commonly used in OOD detection.

Another common approach is to generate OOD samples for training network models. GAN\cite{goodfellow2020generative} can be used to produce low-confidence samples for training networks, as seen in the works by \cite{lee2017training,vernekar2019out,sricharan2018building}. 
Du et al.\cite{du2024dream} use diffusion model to sample outliers in the low-confidence region of ID data to generate OOD samples. Adversarial samples are also used as OOD data to train classifiers\cite{liang2017enhancing,hsu2020generalized}.

\begin{figure*}
    \centering
    \includegraphics[scale=0.65]{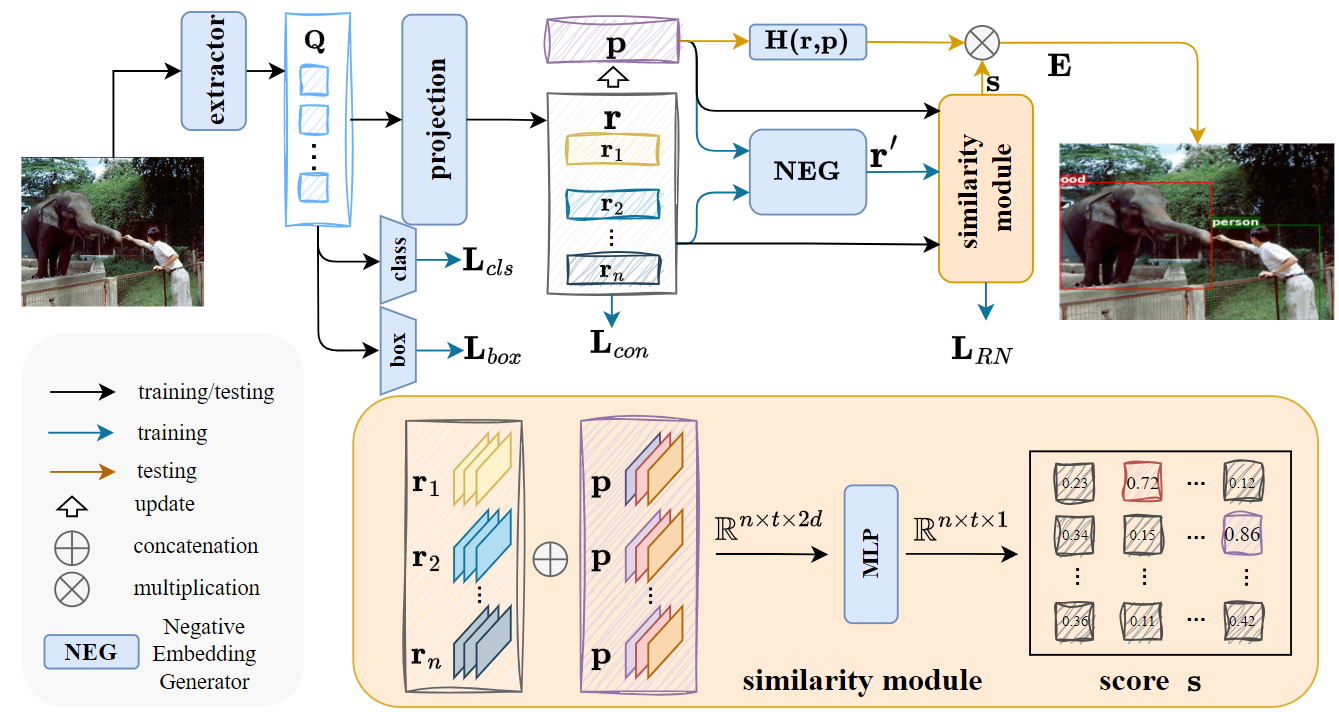}
    \caption{\textbf{The framework of Proto-OOD}. During the training stage, Proto-OOD collects category prototypes from embeddings \textbf{r}. The negative embedding generator(NEG) creates negative embeddings $\textbf{r}'$ by \textbf{r} and \textbf{p}. In the testing stage, \textbf{p} and the similarity module's outputs \textbf{s} are used to calculate \textbf{E} to determine if an object is OOD object. \textbf{H(r,p)} outputs the cosine similarity between the embedding \textbf{r} and the prototypes \textbf{p}.
}
\end{figure*}

Some methods detect OOD data by monitoring the distribution differences in neural network outputs. Yu et al.\cite{yu2019unsupervised} add an auxiliary classifier to the model to identify OOD data by comparing the output differences between the two classifiers. 
Wang et al.\cite{wang2022vim} generate a virtual logit for OOD data. By comparing the output logit with the virtual logit, they can detect OOD data.

Many research methods achieve good OOD detection in classification tasks, but their performance in the more complex field of object detection is still not satisfactory.

\subsection{OOD Detection for Object Detection}
Research methods for OOD object detection are still relatively scarce. 
SIREN \cite{du2022siren} proposes a distance-based method that maps features onto a hypersphere surface. It identifies OOD data by assigning higher score to OOD samples via trainable distance parameters.
However, parameters trained on ID data are insufficient for distinguishing OOD data because SIREN\cite{du2022siren} cannot learn OOD features during the training stage. 
VOS\cite{du2022vos} samples outliers from feature space to generate OOD samples for training a classifier. 
VOS\cite{du2022vos} uses Mahalanobis distance to limit the sampling scope, which may misclassify boundary ID data as OOD. SAFE\cite{Wilson_2023_ICCV} uses FGSM\cite{goodfellow2014explaining} to generate adversarial samples as OOD data for training networks, achieving strong detection performance. However, the generated adversarial samples have overly uniform features, leading to a significant drop in performance when ID and OOD data features are dissimilar.

\subsection{Prototypes for Few-Shot Learning}
Prototypes are widely used in few-shot learning. Prototype network\cite{snell2017prototypical} collects the means of support set features as prototypes and compute the distance from the input features to the prototypes to determine the category. 
Liu et al.\cite{DBLP:journals/corr/abs-1911-10714} propose a prototype network based on cosine similarity to compute prototypes for few-shot learning. 
Zhao et al.\cite{zhao2022fs3d} employ prototypes for few-shot learning on point cloud object detection. 
In their method, they use a momentum coefficient to update the prototypes during the training stage, making the prototypes more robust. 
Gao et al.\cite{pmlr-v161-gao21a} propose contrastive prototype learning (CPL) for few-shot learning. 
CPL aims to pull the query samples of the same category features closer and those of different categories further away. 
Allen et al.\cite{pmlr-v97-allen19b} propose infinite mixture prototypes to adaptively represent data distributions.

\section{Method}
In this section, we present our approach to OOD object detection. The preliminaries and overviews are introduced in sections A and B. In section C, we elucidate the prototype learning and contrastive loss. Following that, in section D, the similarity module and the negative embedding generator are introduced. In section E, we detail how to train Proto-OOD and how to conduct OOD object detection during the testing stage.
\subsection{Preliminaries}
Let the input image set and label set be \textbf{X} and \textbf{Y}, respectively. $Y_{id}$=\{1,2,..,t\} denotes ID data category space. $Y_{ood}$ denotes the OOD data category space ($Y_{id} \cap Y_{ood}=\emptyset$ ). For each $\textbf{y} \in$ $\textbf{Y}$, $\textbf{y}$ = \{$l,x_{min},y_{min},x_{max},y_{max}$\} where $l$ and $(x_{min},y_{min},x_{max},y_{max})$ denote the category label and the coordinate of the object, respectively. During the training stage, the ID dataset $\textbf{D}^{ID}$($l \in Y_{id}$) is used to train the model's parameters. A subset of the ID dataset $\textbf{D}^{ID}$ and the OOD dataset $\textbf{D}^{OOD}$($l \in Y_{ood}$) are utilized to compute evaluation metrics during evaluation. 

During the testing stage, the model not only detects the category and location of the object, but also outputs \textbf{E} to distinguishes whether the object belongs to ID or OOD data. If \textbf{E} exceeds a hyperparameter $\gamma$, the object is detected as in-distribution; otherwise, it is detected as out-of-distribution.

\subsection{Overviews}
We illustrate the framework in Fig. 2. The network includes a feature extractor, similarity module, project head, and negative embedding generator. Given an input image \textbf{x}, the feature extractor, utilizing a transformer architecture, processes the image to extract feature query \textbf{Q}. The query \textbf{Q} is used to predict the object category and box using the class head and the box head. The project head is a two-layer MLP that maps the query \textbf{Q} to a lower-dimensional embedding \textbf{r}. Subsequently, the similarity module utilizes the embedding \textbf{r} and the prototypes \textbf{p} to predict the similarity score \textbf{s}. The similarity score \textbf{s} and the cosine similarity between the embedding \textbf{r} and the prototypes \textbf{p} are utilized to calculate the \textbf{E}, which is used to determine whether the object belongs to ID or OOD data. The details of our algorithm are summarized in Algorithm 1.

\subsection{Prototype Learning and Contrastive Loss}
In the prototype network\cite{snell2017prototypical}, prototypes are used to determine the category of input features. In Proto-OOD, the similarity between input features and prototypes is calculated to determine whether the input features belong to OOD data or ID data. Proto-OOD uses the project head to map the query $\textbf{Q} \in \mathbb{R}^{n \times h}$ to a low-dimensional embedding $\textbf{r} \in \mathbb{R}^{n \times d}$ $(d<h)$, and collects prototypes from the embedding \textbf{r}. In the early stage of training, the embedding \textbf{r} may not effectively represent category features. Therefore, Proto-OOD starts collecting prototypes after $\lambda$ epochs, using the following method:

\begin{align}
   \textbf{p}_c  = (\alpha\textbf{p}_c+(1-\alpha)\textbf{r}_c)
\end{align}
where $\alpha$ is the update factor and $\textbf{r}_c$ refers to embeddings from category $c$. The original value of $\textbf{p}_c$ is $\textbf{0}$.

To increase the distinctiveness of the prototypes, contrastive loss is added during the training stage. The contrastive loss $\textbf{L}_{con}$ is calculated as follows:
\begin{align}
    \textbf{L}_{con}=\frac{1}{M}\cdot\sum_{i=1}^{M}f(z_i)
\end{align}

\begin{align}
    f(z_i)=-log\frac{\sum_{j=1,j\neq i}^{M}\left \{{l_i==l_j} \right \}\cdot exp( z_i^\mathsf{T}z_j/\tau))}{\sum_{j=1,j\neq i}^{M}exp( z_i^\mathsf{T}z_j/\tau)}
\end{align}

where  $z_i$=$\frac{r_i}{\|r_i\|}$. $\|r_i\|$ denotes the norm of $r_i$. $\tau$ is a scaling factor. $l_i$ represents the category label of ${r}_i$. $\|l_i\|$ denotes the number of objects of category $l_i$. $M$ is the total number of objects.
\subsection{Similarity Module and Negative Embedding Generator}
After collecting representative prototypes, the similarity between input features and prototypes can be used to determine if an object is OOD data. Inspired by the relation network\cite{sung2018learning}, Proto-OOD uses a similarity module to calculate the similarity between embeddings \textbf{r} and prototypes \textbf{p}. The relation network\cite{sung2018learning}, proposed for few-shot learning, uses a two-layer MLP to calculate the similarity between support set features and query set features to determine the category of query features. Proto-OOD takes embeddings $\textbf{r} \in \mathbb{R}^{n \times d}$ as input for the similarity module, and concatenates \textbf{r} with prototypes $\textbf{p} \in \mathbb{R}^{t \times d}$ to form a fused feature \textbf{f} $\in \mathbb{R}^{n \times t \times 2d}$, where t is the number of categories. The similarity score \textbf{s} is predicted by a two-layer MLP.
\begin{align}
    \textbf{s} = Sigmoid(\sigma(\sigma(\textbf{f})))
\end{align}

where $\sigma$ denotes ReLU(Linear($\cdot$)), and \  $\textbf{s}$ $\in$ $\mathbb{R}^{n \times t \times 1}$. 

\begin{figure}[htp]
    \includegraphics[scale=0.5]{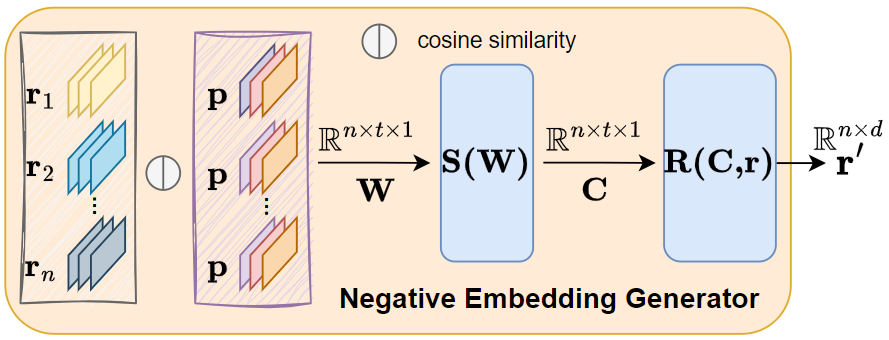}
    \centering
    \caption{The details of negative embedding generator.}
    \label{fig:proto-vis}
\end{figure}

The negative embedding generator, as shown in Fig. 3, first calculates the cosine similarity $\textbf{W} \in \mathbb{R}^{n \times t \times 1}$ between the embedding \textbf{r} and prototypes \textbf{p}. Then, it generates a weight matrix $\textbf{C} \in \mathbb{R}^{n \times t \times 1}$ using Equation(5) to adjust the weights among the different category prototypes. A scaling factor \textbf{T} is used to balance the weights of different categories.

% \begin{align}
%     \textbf{C}=\textbf{S(W)}=\textbf{Softmax((1-\textbf{W}))/T)}
% \end{align}
\begin{align}
   \textbf{C}=\textbf{S(W)}=\textbf{Softmax((1-\textbf{W}))/T)}
\end{align}
We use Equation(6) to generate the negative embeddings $\textbf{r}'$ $\in \mathbb{R}^{n\times d} $. 
\begin{align}
    \textbf{r}'=\textbf{R(C,r)}
\end{align}

where $\textbf{R(C,r)}$ process the input as follows :
\begin{align}
\textbf{r}'_i=\textbf{r}_i+\sum_{j=0}^t(\textbf{C}_{ij}\cdot\textbf{p}_j)
\end{align}
for each $\textbf{r}_i\in \textbf{r}$, it generates a negative embedding $\textbf{r}'_i \in \textbf{r}'$. The $t$ in Equation(7) denotes the number of categories.

% Negative embeddings $\textbf{r}'$ and background features from images are used as negative samples to train the similarity module. During the loss calculation phase, the output embeddings of the Hungarian algorithm are used as positive samples.

\subsection{Training and Testing}
Training and testing procedures are represented in Algorithm 1. During the training stage, because the representation of embeddings \textbf{r} is poor in the early stage, Proto-OOD sets a hyperparameter $\lambda$ to determine when to start collecting prototypes. Specifically, when the training epoch is greater than $\lambda$, Proto-OOD begins to collect prototypes.

We first train the object detector with parameters $\theta$ and the project head with parameters $\phi$ in stage1 (3: in Algorithm 1). The loss function is as follows:
\begin{align}
    \textbf{L}_{stage1}=\textbf{L}_{cls}+\textbf{L}_{box}+\textbf{L}_{con}
\end{align}

where $\textbf{L}_{cls}$ and $\textbf{L}_{box}$ refer to the object detection loss and $\textbf{L}_{con}$ refers to the contrastive loss from Equation(2).

When the epoch is greater than $\lambda$, Proto-OOD begins to collect prototypes. After $\omega$ epochs, Proto-OOD begins to train the  similarity module. The loss in stage2 (7: in Algorithm 1) as follows:
\begin{align}
    \textbf{L}_{stage2}=\textbf{L}_{cls}+\textbf{L}_{box}+\textbf{L}_{con}+
    \textbf{L}_{RN}
\end{align}

where $\textbf{L}_{RN}$ is the focal loss\cite{lin2017focal}, and its formula is as follows:
\begin{align}
    \textbf{L}_{RN}=\left\{
\begin{aligned}
- ( 1 - p )^\gamma \log ( p ),\quad \quad 
if\quad y=1\\
-p^\gamma \log(1-p),\quad otherwise
\end{aligned}
\right.
\end{align}

where $p$ represents the predicted probability, and $y$ represents the label.

During the testing stage, the object detector not only predicts the category score and box position, but also outputs the object feature embeddings \textbf{r} and similarity score \textbf{s}. The \textbf{E}, calculated from $\textbf{r}$ and prototypes $\textbf{p}$, is used to determine if the input is an OOD object. The $\textbf{E}$ is calculated as follows:
\begin{align}
    \textbf{E}= \textbf{H(}\textbf{r},\textbf{p)}\cdot \textbf{s}=\textbf{exp(}\textbf{Cosine\_similarity(}\textbf{r},\textbf{p}\textbf{))}\cdot \textbf{s}
\end{align}

where the \textbf{H(r,p)} outputs the cosine similarity between the embedding \textbf{r} and the prototypes \textbf{p}. The \textbf{exp(·)} denotes the exponential operation.
\begin{algorithm}
	\renewcommand{\algorithmicrequire}{\textbf{Input:}}
	\renewcommand{\algorithmicensure}{\textbf{Output:}}
	\caption{training and testing}
	\label{alg:1}
	\begin{algorithmic}[1]
		\REQUIRE ID dataset $\textbf{D}^{IN}$, randomly initialize object detector with parameters $\theta$, project head with parameters $\phi$, similarity module with parameters $\psi$, $\lambda$, $\omega$.
		\ENSURE object detector, project head, similarity module, prototypes.
            \WHILE{training}
            \IF{epoch$<$$\lambda$}
            \STATE Update object detector parameters $\theta$, project head parameters $\phi$ by Equation(8).
            \ELSIF{$\lambda$$<$epoch$<$$(\lambda + \omega)$}
            \STATE Update object detector parameters $\theta$, project head parameters $\phi$ by Equation(8).
            Collect prototypes by Equiation (1).
            \ELSE 
            \STATE Update object detector parameters $\theta$, project head parameters $\phi$, similarity module parameters $\psi$ by Equiation (9). Collect prototypes by Equiation (1).
            \ENDIF
            \ENDWHILE
		\WHILE{testing}
            \STATE Detect objects by object detector with prototypes, and output class score, box location, embeddings $\textbf{r}$, similarity score $\textbf{s}$.
            \STATE Calculate the $\textbf{E}$ by Equation(11). Output OOD detection results.
            \ENDWHILE

	\end{algorithmic}  
\end{algorithm}

We set a threshold $\gamma$ for the $\textbf{E}$ to distinguish between ID and OOD objects:
\begin{equation}
\textbf{g}=\left\{
\begin{aligned}
& 1,\quad \quad \textbf{E}>=\gamma\\
& 0,\quad \quad \textbf{E}<\gamma
\end{aligned}
\right.
\end{equation}

if $\textbf{g}$ equals 1, the object is considered ID data. If $\textbf{g}$ equals 0, the object is considered OOD data.

\section{Experiments}
\begin{table*}
    \small %
  \caption {Detection results comparing our method to current OOD object detection method.
  }
\renewcommand\arraystretch{1.2}
\resizebox{0.98\linewidth}{!}{%

\begin{tabular}{c|c|cc|cc|c}
\hline
\multirow{2}{*}{\textbf{ID Dataset}} & \multirow{2}{*}{\textbf{Method}}       & \multicolumn{2}{c|}{\textbf{FPR95}}      & \multicolumn{2}{c|}{\textbf{AUROC}}             & \multirow{2}{*}{\textbf{mAP}} \\
                                     &                                        & \textbf{MS-COCO}   & \textbf{OpenImages} & \textbf{MS-COCO}   & \textbf{OpenImages}        &                               \\ \hline
\multirow{7}{*}{Pascal VOC}          & SIREN-VMF\cite{du2022siren}            & 75.49±0.8          & 78.36±1.0           & 76.10±0.1          & 71.05±0.1                  & 60.8±0.1                      \\
                                     & SIREN-KNN\cite{du2022siren}            & 64.77±0.2          & 65.99±0.5           & 78.23±0.2          & \hspace{0.65em}74.93±0.1 ~ & 60.8±0.1                      \\
                                     & VOS-RegNetX4.0\cite{du2022vos}         & 47.77±1.1          & 48.33±1.6           & 89.00±0.4          & 87.59±0.2                  & 51.6±0.1                      \\
                                     & VOS-ResNet50\cite{du2022vos}           & 47.53±2.9          & 51.33±1.6           & 88.70±1.2          & 85.23±0.6                  & 48.9±0.2                      \\
                                     & SAFE-ResNet50\cite{Wilson_2023_ICCV}   & 47.40±3.8          & 20.06±2.3           & 80.30±2.4          & 92.28±1.0                  & -                             \\
                                     & SAFE-RegNetX4.0\cite{Wilson_2023_ICCV} & 36.32±1.1          & \textbf{17.69±1.0}  & 87.03±0.5          & \textbf{94.30±0.2}         & -                             \\ \cline{2-7} 
                                     & Proto-OOD(ours)                        & \textbf{30.98±2.5} & 33.25±3.7           & \textbf{90.23±0.2} & 89.73±1.0                  & \textbf{65.4±0.2}             \\ \hline
\multirow{7}{*}{BDD100k}             & SIREN-VMF\cite{du2022siren}            & 67.54±1.3          & 66.31±0.9           & 80.06±0.5          & 79.77±1.2                  & 31.3±0.0                      \\
                                     & SIREN-KNN\cite{du2022siren}            & 53.97±0.7          & 66.31±0.9           & 86.56±0.1          & 89.00±0.4                  & 31.3±0.0                      \\
                                     & VOS-RegNetX4.0\cite{du2022vos}         & 36.61±0.9          & 27.24±1.3           & 89.08±0.6          & 89.00±0.4                  & \textbf{32.5±0.1}             \\
                                     & VOS-ResNet50\cite{du2022vos}           & 44.27±2.0          & 35.54±1.7           & 86.87±2.1          & 88.52±1.3                  & 31.3±0.0                      \\
                                     & SAFE-ResNet50\cite{Wilson_2023_ICCV}   & 32.56±0.8          & 16.04±0.5           & 88.96±0.6          & 94.64±0.3                  & -                             \\
                                     & SAFE-RegNetX4.0\cite{Wilson_2023_ICCV} & 21.69±0.5          & \textbf{13.98±0.3}  & 93.91±0.1          & \textbf{95.57±0.1}         & -                             \\ \cline{2-7} 
                                     & Proto-OOD(ours)                    & \textbf{19.78±1.7} & 15.61±2.2           & \textbf{94.05±0.2} & 93.17±0.1                  & 31.7±0.1                      \\ \hline
\end{tabular}
}
\end{table*}
In Section A, we present the experimental setup. In Section B, we introduce the evaluation metrics and come up with an improved evaluation protocol. Experimental results are also displayed in this section. The results of ablation experiments and quantitative experiments are presented in the following sections.

\subsection{Experimental Settings}
\textbf{Datasets.} Same as SIREN\cite{du2022siren}, we use the PASCAL-VOC\cite{Everingham15} and Berkeley DeepDrive-100K\cite{DBLP:journals/corr/XuGYD16} (BDD100K) datasets as ID datasets for training and evaluation. The subsets of the MS-COCO\cite{DBLP:journals/corr/LinMBHPRDZ14} and OpenImages\cite{OpenImages} are used as OOD datasets for evaluation.

\textbf{Implementation Details.} 
The RT-DETR\cite{lv2023detrs}, with ResNet50\cite{he2016deep} as its backbone, is utilized as an object detector. We set the scaling factor $\tau$ = 0.2 in Equation(3), and the \textbf{T} = 2 in Equation(5). For the Pascal VOC\cite{Everingham15} dataset, the $\lambda$ = 40 and the $\omega$ = 5 in Algorithm 1. For the BDD100K\cite{DBLP:journals/corr/XuGYD16}, the $\lambda$ = 25 and the $\omega$ = 5. For OOD metrics evaluation, we selectively choose K objects with the highest scores predicted by the class head for an image when evaluating the ID dataset, where K corresponds to the number of annotated objects in an image.

\subsection{Evaluation Metrics and Results}
$\textbf{Protocol}_A$ 
We utilize the evaluation protocol defined by SIREN\cite{du2022siren} as $\textbf{Protocol}_A$ to evaluate performance. We report the metric mAP for the ID dataset. For the OOD dataset, we report the false positive rate of samples when the true positive rate of ID objects is at 95\% (FPR95) and the area under the receiver operating characteristic curve (AUROC). Lower FPR95 means the model is less likely to misclassify OOD objects as ID objects, while higher AUROC indicates better overall performance on both ID and OOD datasets.

Additionally, we note that previous methods\cite{du2022siren,du2022vos,Wilson_2023_ICCV} do not filter inaccurate predictions when calculating metrics, which may lead to inaccurate OOD assessment metrics. Specifically, the formula for False Positive Rate (FPR) is as follows:
\begin{align}
   FPR  = \frac{FP}{FP+TN}
\end{align}
where $FP$ represents the misclassification of negative samples as positive samples and $TN$ indicates that the negative sample is correctly predicted to be negative. 
Previous methods\cite{du2022siren,du2022vos,Wilson_2023_ICCV} do not filter out incorrect predictions when calculating FPR, leading to an overestimation of $FP$ in Equation(13). Consequently, this leads to inaccurate calculations of the $FPR$.

$\textbf{protocol}_B$ We propose a new evaluation protocol, $\textbf{protocol}_B$. To avoid inaccuracies that lead to an increase in false positives (FP), only the top K objects with the highest scores in the ID dataset are selected for calculation, where K equals the number of annotated objects in the image.

\begin{figure*}
    \centering
    \includegraphics[scale=0.6]{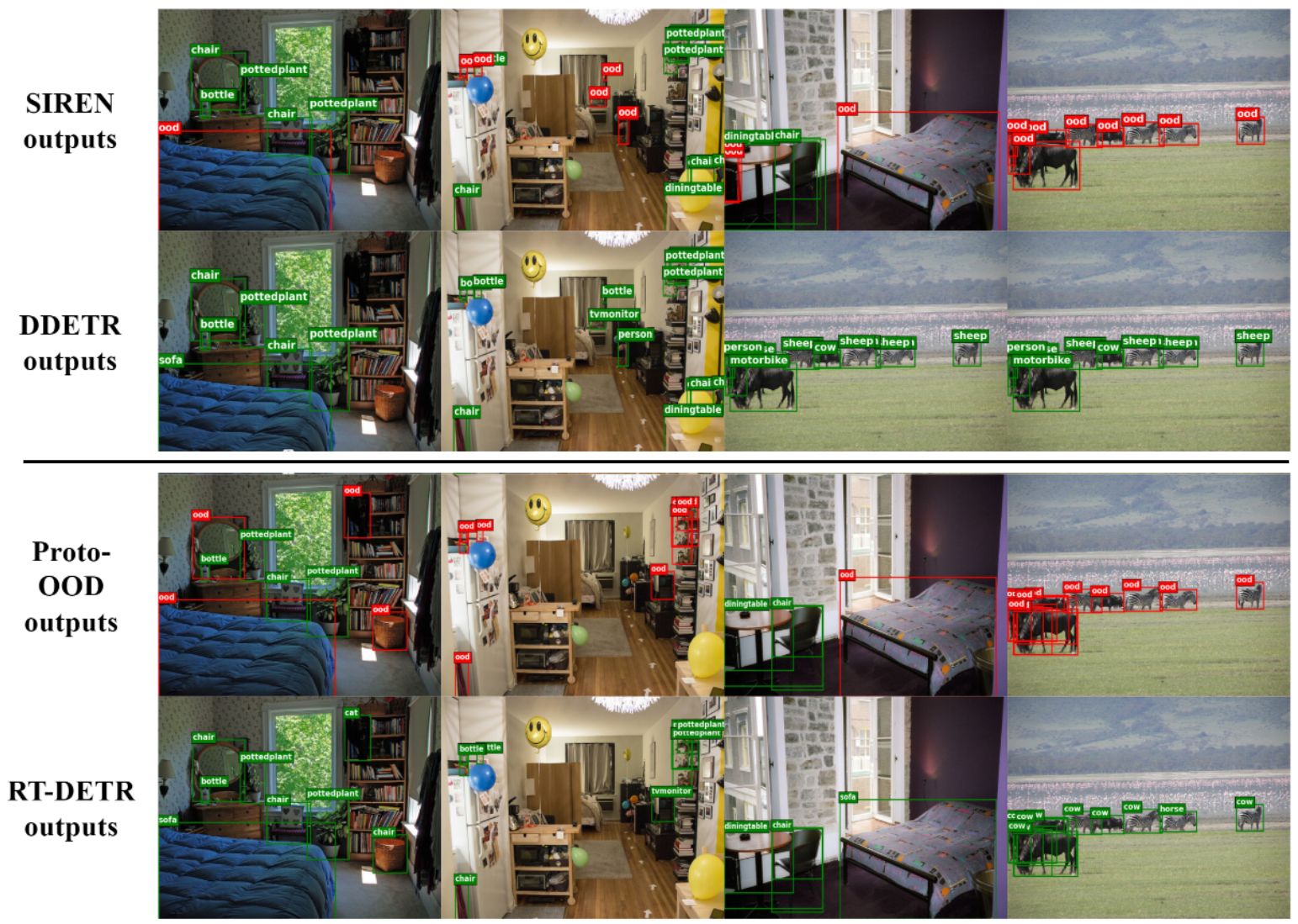}
    \caption{Visualization the outputs of two object detection networks, RT-DETR\cite{lv2023detrs} and DDETR\cite{zhu2020deformable}, on OOD dataset images, as well as the detection outputs of Proto-OOD on RE-DETR\cite{lv2023detrs} and SIREN\cite{du2022siren} applied to DDETR \cite{zhu2020deformable}.
}
\end{figure*}

\textbf{Results and Analysis.} 
The performance of Proto-OOD under $\textbf{protocol}_B$ compared with other methods is shown in Table I. When the Pascal VOC\cite{Everingham15} dataset is used as the ID dataset, Proto-OOD achieves the best FPR95 and AUROC on MS-COCO\cite{DBLP:journals/corr/LinMBHPRDZ14}. When the BDD100K\cite{DBLP:journals/corr/XuGYD16} dataset is used as the ID dataset, Proto-OOD also achieves the  promising results. We analyzed three possible reasons:

(1) BDD100K\cite{DBLP:journals/corr/XuGYD16} has a large number of images in the training set, enabling Proto-OOD to collect highly representative prototypes. Also, as a road scene dataset, BDD100K\cite{DBLP:journals/corr/XuGYD16} differs significantly from the scenes in MS-COCO\cite{DBLP:journals/corr/LinMBHPRDZ14} and OpenImages\cite{OpenImages}. Thus, Proto-OOD achieves good detection results on MS-COCO\cite{DBLP:journals/corr/LinMBHPRDZ14} and OpenImages\cite{OpenImages}.

(2) BDD100K\cite{DBLP:journals/corr/XuGYD16} has significantly more images in the test set compared to MS-COCO\cite{DBLP:journals/corr/LinMBHPRDZ14} and OpenImages\cite{OpenImages}, thus having a greater impact on the evaluation metrics.

We evaluate all methods in $\textbf{protocol}_B$, with results shown in Table II. The results indicate that previous methods\cite{du2022siren,du2022vos,Wilson_2023_ICCV} can accurately reflect the model's OOD detection capability when inaccurate outputs are filtered. The FPR95 for SIREN\cite{du2022siren} decreases by more than 30\%. The performance of VOS\cite{du2022vos} also improves significantly. SAFE continues to get the best performance on OpenImages\cite{OpenImages}.

\begin{table*}
    \small %
  \caption {The results of ablation study. A represents the contrastive loss. B represents the negative embedding generator.
  }
\renewcommand\arraystretch{1.2}
\resizebox{1\linewidth}{!}{%
\begin{tabular}{c|c|cc|cc}
\hline
\multirow{2}{*}{\textbf{ID Dataset}} & \multirow{2}{*}{\textbf{Method}} & \multicolumn{2}{c|}{\textbf{FPR95}}    & \multicolumn{2}{c}{\textbf{AUROC}}     \\
                                     &                                  & \textbf{MS-COCO} & \textbf{Openimages} & \textbf{MS-COCO} & \textbf{Openimages} \\ \hline
\multirow{3}{*}{Pascal VOC}          & Proto-OOD w/o A,B                & 35.63            & 37.16               & 87.92            & 86.43               \\
                                     & Proto-OOD w/o B                  & 31.84            & 34.87               & 89.11            & 88.04               \\
                                     & Proto-OOD                        & \textbf{30.98}   & \textbf{33.25}      & \textbf{90.23}   & \textbf{89.73}      \\ \hline
\end{tabular}
}
\end{table*}

\subsection{Ablation Study}
We conduct ablation studies to assess the impact of the contrastive loss and the negative embedding generator. Without contrastive loss and the negative embedding generator, Proto-OOD only uses background features as negative examples to train the similarity module. From Table II, it shows that Proto-OOD's detection performance improved after adding the contrastive loss and the negative embedding generator.
\begin{figure}
    \includegraphics[scale=0.3]{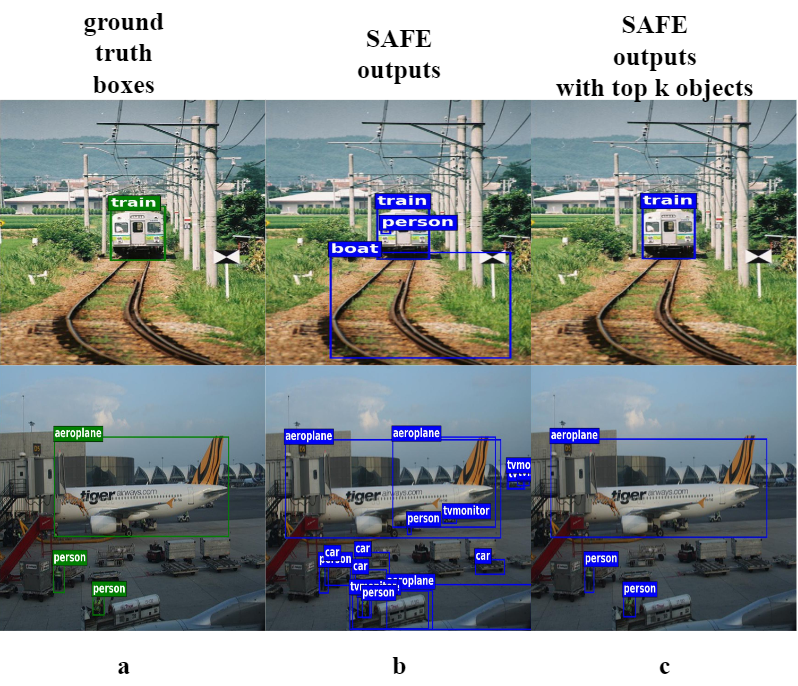}
    \centering
    \caption{\textbf{b} displays the predictions of the SAFE object detector. All the predictions are used to calculate matrix under $\textbf{protocol}_A$, including the incorrect predictions. \textbf{c} shows the predictions, which are used to calculate matrix under $\textbf{protocol}_B$.
}
\end{figure}

\subsection{Qualitative Analysis}
During the testing stage, a threshold $\gamma$ in Equation(12) is used to distinguish between ID and OOD data. Fig. 4 shows the visualization results of different models on images from MS-COCO. 

Using Pascal VOC as the ID dataset, Fig. 4 displays the prediction results of two object detection networks, RT-DETR\cite{lv2023detrs} and DDETR\cite{zhu2020deformable}, on OOD dataset images, as well as the detection results of Proto-OOD on RE-DETR\cite{lv2023detrs} and SIREN\cite{du2022siren} applied to DDETR\cite{zhu2020deformable}.

\subsection{Limitation of Current Evaluation Metrics}
We discuss the limitations of the current evaluation metrics as follows:
\begin{table*}
    \small %
  \caption {Detection results comparing our method to current OOD object detection method. $^*$: results tested under the $\textbf{protocol}_B$.
  }
\renewcommand\arraystretch{1.2}
\resizebox{1\linewidth}{!}{%
\begin{tabular}{c|c|cc|cc|c}
\hline
\multirow{2}{*}{{\textbf{ID Dataset}}} & \multirow{2}{*}{{\textbf{Method}}}       & \multicolumn{2}{c|}{\textbf{FPR95}}    & \multicolumn{2}{c|}{\textbf{AUROC}}      & \multirow{2}{*}{{\textbf{mAP}}} \\
                                       &                                          & \textbf{MS-COCO} & \textbf{Openimages} & \textbf{MS-COCO} & \textbf{Openimages}   &                                 \\ \hline
\multirow{7}{*}{{Pascal VOC}}          & SIREN-VMF\cite{du2022siren}              & 75.49            & 78.36               & 76.10            & 76.10                 & 60.8                            \\
                                       & SIREN-VMF$^*$\cite{du2022siren}          & 44.60            & 54.14               & \textbf{90.95}   & \hspace{0.65em}85.39 ~ & 60.8                            \\
                                       & VOS-ResNet50\cite{du2022vos}             & 47.53            & 51.33               & 89.00            & 87.59                 & 51.6                            \\
                                       & VOS-ResNet50$^*$\cite{du2022vos}         & 35.64            & 44.16               & 90.06            & 85.23                 & 48.9                            \\
                                       & SAFE-ResNet50\cite{Wilson_2023_ICCV}     & 47.40            & 20.06               & 80.30            & 92.28                 & -                               \\
                                       & SAFE-ResNet50$^*$\cite{Wilson_2023_ICCV} & 44.74            & \textbf{22.97}      & 82.37            & \textbf{92.61}        & -                               \\ \cline{2-7} 
                                       & Proto-OOD$^*$(ours)                      & \textbf{30.98}   & 33.25               & 90.23            & 89.73                 & \textbf{65.4}                   \\ \hline
\end{tabular}
}
\end{table*}

(1) Previous papers\cite{du2022siren,du2022vos,Wilson_2023_ICCV} do not filter out false positives with background in the boxes when using the ID dataset to calculate OOD evaluation metrics. SAFE's object detector predicts the rail as a boat and uses it as ID data to participate in the calculation of the OOD evaluation metrics (in Fig. 5 \textbf{b}). To correct this error, $\textbf{Protocol}_B$ selects K objects with the highest classification scores from the class head's outputs to participate in the OOD evaluation metrics calculation, where K equals the number of objects annotated in an image. Fig. 5 \textbf{c} shows the results after processing. Table III, comparing results under $\textbf{Protocol}_A$ and $\textbf{Protocol}_B$, indicates that $\textbf{Protocol}_B$ successfully minimizes the impact of erroneous predictions in the calculation of the OOD evaluation metrics.

(2) The imbalance between the number of images in the ID dataset and the OOD dataset leads to an inaccurate evaluation of the model's performance. BDD100K is used as the ID dataset, and 10,000 images are designated as ID data to calculate OOD evaluation metrics, whereas 930 images from MS-COCO are used as the OOD data to calculate OOD evaluation metrics. This imbalance results in evaluation metrics being more affected by the BDD100K dataset.

\section{Conclusion}
In this paper, we propose a novel framework, the Proto-OOD, which utilizes prototypes to detect OOD objects. Proto-OOD gradually collects prototypes of ID data during the training stage, and generates negative samples in the feature space to train a similarity module. In addition, we propose a more reasonable OOD evaluation protocol. We hope that our prototype-based approach can inspire further research on OOD object detection in the future.
\bibliographystyle{IEEEtran}
\bibliography{IEEEabrv,ref}

\end{document}